\documentclass{article}
\usepackage{arxiv}

\usepackage[utf8]{inputenc} % allow utf-8 input
\usepackage[T1]{fontenc}    % use 8-bit T1 fonts
\usepackage{url}            % simple URL typesetting
\usepackage{booktabs}       % professional-quality tables
\usepackage{amsfonts}       % blackboard math symbols
\usepackage{nicefrac}       % compact symbols for 1/2, etc.
\usepackage{graphicx}
\usepackage{doi}
\usepackage{balance}
\usepackage{graphicx}
\usepackage{algorithm}
\usepackage{color}
\usepackage{siunitx}
\usepackage{etoolbox}
\usepackage{tabularx}
\usepackage{adjustbox}
\usepackage{array}
\usepackage{xparse}
\usepackage{algpseudocode}
\usepackage{caption}
\usepackage{subcaption}
\usepackage{glossaries}
\usepackage{amsmath,amsfonts}%,amssymb,amsfonts,mathtools}
\usepackage{bm}

\usepackage[backend=biber,style=LNI,natbib=true,giveninits=false]{biblatex}

\NewDocumentCommand{\rot}{O{30} O{1em} m}{\makebox[#2][l]{\rotatebox{#1}{#3}}}%

\newcommand{\batchSize}{b}

\newcommand{\learner}{\theta}

\newcommand{\labeledSet}{\mathcal{L}}

\newcommand{\unlabeledSet}{\mathcal{U}}

\newcommand{\unlabeledQuery}{Q}

\newcommand{\singleSample}{x}

\newcommand{\ALStrategy}{f}
\newcommand{\LC}{\ALStrategy_{LC}}

\newcommand{\conf}{P}
\newcommand{\softmax}{\sigma}
\newcommand{\logit}{z}

\newcommand{\dist}{dist}

\DeclareMathOperator*{\argmax}{argmax}

\newacronym{ML}{ML}{Machine Learning}
\newacronym{NLP}{NLP}{Natural Language Processing}
\newacronym{CNN}{CNN}{Convolutional Neural Networks}
\newacronym{AL}{AL}{Active Learning}
\newacronym{RL}{RL}{Reinforcement Learning}
\newacronym{MDP}{MDP}{Markov Decision Problem}
\newacronym{NN}{NN}{Neural Network}
\newacronym{LC}{LC}{Uncertainty Least Confidence}
\newacronym{Rand}{Rand}{Random Sampling}
\newacronym{MM}{MM}{Uncertainty Max-Margin}
\newacronym{Ent}{Ent}{Uncertainty Entropy}
\newacronym{QBC}{QBC}{Query-by-committee}
\newacronym{KLD}{KLD}{Kullback-Leibler Divergence}
\newacronym{VE}{VE}{Vote Entropy}
\newacronym{UC}{UC}{Uncertainty Clipping}
\newacronym{Evi}{Evi}{Evidential Neural Networks}
\newacronym{IS}{IS}{Inhibited Softmax}
\newacronym{LS}{LS}{Label Smoothing}
\newacronym{MC}{MC}{Monte-Carlo Dropout}
\newacronym{TeSc}{TeSc}{Temperature Scaling}
\newacronym{TrSc}{TrSc}{TrustScore}
\glsdisablehyper
\newcolumntype{C}[1]{>{\centering\arraybackslash}p{#1}}
\newcolumntype{R}[1]{>{\raggedleft\arraybackslash}p{#1}}
\newcolumntype{L}[1]{>{\raggedright\arraybackslash}p{#1}}

\MakeRobust{\Call}

\title{To Softmax, or not to Softmax: that is the question when applying Active Learning for Transformer Models}
\date{October 06, 2022}

\author{Julius Gonsior\\
	Technische Universität Dresden\\
	Dresden, Germany\\
	\texttt{julius.gonsior@tu-dresden.de} \\
	\And
	Christian Falkenberg\\
	Technische Universität Dresden\\
	Dresden, Germany\\
	\texttt{christian.falkenberg.@tu-dresden.de} \\
	\And
	Silvio Magino\\
	Technische Universität Dresden\\
	Dresden, Germany\\
	\texttt{silvio.magino@tu-dresden.de} \\
	\And
	Anja Reusch\\
	Technische Universität Dresden\\
	Dresden, Germany\\
	\texttt{anja.reusch@tu-dresden.de} \\
	\And
	Maik Thiele \\
	Hochschule für Technik und Wirtschaft Dresden\\
	Dresden, Germany\\
	\texttt{maik.thiele@htw-dresden.de} \\
	\And
	Wolfgang Lehner\\
	Technische Universität Dresden\\
	Dresden, Germany\\
	\texttt{wolfgang.lehner@tu-dresden.de} \\
}

\renewcommand{\shorttitle}{To Softmax, or not to Softmax}

\hypersetup{
pdftitle={To Softmax, or not to Softmax: that is the question when applying Active Learning for Transformer Models},
pdfsubject={q-cs.LG, q-cs.AI},
pdfauthor={Julius Gonsior, Christian Falkenberg, Silvio Magino, Anja Reusch, Maik Thiele, Wolfgang Lehner},
pdfkeywords={Active Learning, Transformer, Softmax, Uncertainty, Calibration, Deep Neural Networks},
}

\begin{document}

\maketitle

\begin{abstract}
%This is a brief overview of the paper, which should be 70 to 150 words long and include the most relevant points. This has to be a single paragraph.
Despite achieving state-of-the-art results in nearly all Natural Language Processing applications, fine-tuning Transformer-based language models still requires a significant amount of labeled data to work. A well known technique to reduce the amount of human effort in acquiring a labeled dataset is \textit{Active Learning} (AL): an iterative process in which only the minimal amount of samples is labeled. AL strategies require access to a quantified confidence measure of the model predictions. A common choice is the softmax activation function for the final layer. As the softmax function provides misleading probabilities, this paper compares eight alternatives on seven datasets. Our almost paradoxical finding is that most of the methods are too good at identifying the true most uncertain samples (outliers), and that labeling therefore exclusively outliers results in worse performance. As a heuristic we propose to systematically ignore samples, which results in improvements of various methods compared to the softmax function.

\end{abstract}

\section{Introduction}
The most common use case of \gls{ML} is supervised learning, which inherently requires a labeled dataset to demonstrate the desired outcome to the to-be-trained \gls{ML} model. This initial step of acquiring a labeled dataset can only be accomplished by often rare-to-get and costly human domain experts; automation is not possible as the automation via \gls{ML} is exactly the task which should be learned. For example, the average cost for the common label task of segmenting a single image reliably is 6,40 USD\footnote{According to scale.ai as of December 2021}. At the same time, recent advances in the field of \gls{NN} such as Transformer~\cite{transformer} models for \gls{NLP} (with BERT~\cite{BERT} being the most prominent example) or \gls{CNN}~\cite{CNN} for computer vision resulted in huge Deep \gls{NN} which require even more labeled training data. Reducing the amount of labeled data is therefore a primary objective of making \gls{ML} more applicable in real-world scenarios.

%A \textit{transformer}~\cite{transformer} is a deep \gls{NN} model using the self-attention~\cite{transformer} mechanism to process sequential input. Its original use-case was in the field of \gls{NLP}, but it has been applied to other domains as well. The underlying architecture allows to pre-train a Transformer model on a large text corpus in an unsupervised manner. After the pre-training phase the Transformer model has knowledge about natural language and can semantically model text. When applying a Transformer model in a supervised context for classification tasks, the pre-trained model can be extended by a final feed-forward \gls{NN} layer, which then gets trained on a set of labeled data (\textit{finetuning}). The most prominent pre-trained Transformer-encoder model is BERT~\cite{BERT}.

Besides artificially increasing the amount of labeled data by data augmentation, or partially automating the labeling tasks for example through programmatic labeling functions~\cite{Snorkel}, the process of selecting out of the large pool of unlabeled data which samples to label first, called \gls{AL}, is the target to be improved in this work. \gls{AL} is an iterative process, in which in each iteration a new subset of unlabeled samples is selected for labeling using human annotators. 
Due to the iterative selection, the already existent knowledge of the so-far labeled data can be leveraged to select the most promising samples to be labeled next. The goal is to reduce the amount of necessary labeling work while keeping the same model performance. %\gls{AL} achieves this by preventing the annotation of nearly identical, and therefore redundant samples, which the model has already learned. However, the challenge of applying \gls{AL} in this setting is the almost paradoxical problem to be solved: how to decide on a strategy determining which samples are most beneficial to the \gls{ML} model, without knowing the label of the samples, since this is exactly the task to be learned by the to-be-trained \gls{ML} model.
%~\cite{spreadsheet_seadata, galetal17,lowel19}

Despite successful application in a variety of domains, \gls{AL} fails to work for very deep \gls{NN} such as Transformer models, rarely beating pure random sampling. The common explanation~\cite{karamcheti-etal-2021-mind, gleave2022uncertainty, sankararaman2022bayesformer, rejected_openreview} is that \gls{AL} methods favor hard-to-learn samples, often simply called \textit{outliers}, which therefore neglects the potential benefits from \gls{AL}. An additional possible explanation is the potential misuse of the softmax activation function of the final output layer as a method for computing the confidence of the \gls{NN} in its predictions.

Nearly all \gls{AL} strategies rely on a method measuring the certainty of the to-be-trained \gls{ML} model in its prediction as probabilities. The reasoning behind is that those samples with high model uncertainty are the most useful ones to learn from, and should therefore be labeled first. For \gls{NN} typically the \textit{softmax} activation function is used for the last layer, and its output interpreted as the probability of the confidence of the \gls{NN}. But interpreting the softmax function as the true model confidence is a fallacy~\cite{pearce2021understanding}. We compare therefore eight alternative methods for \gls{AL} in an extensive end-to-end evaluation of fine-tuning Transformer models for seven common classification datasets \gls{AL}.

Our main contributions are: 1) comprehensive comparison and evaluation of eight alternative methods to the vanilla softmax function for calculation \gls{NN} model certainty in the context of \gls{AL} for fine-tuning Transformer models, and 2) proposing the novel and easy to implement method \textit{\acrfull{UC}} of mitigating the negative effect of uncertainty based \gls{AL} methods of favoring outliers.

\begin{figure}[!t]
    \centering
    \includegraphics{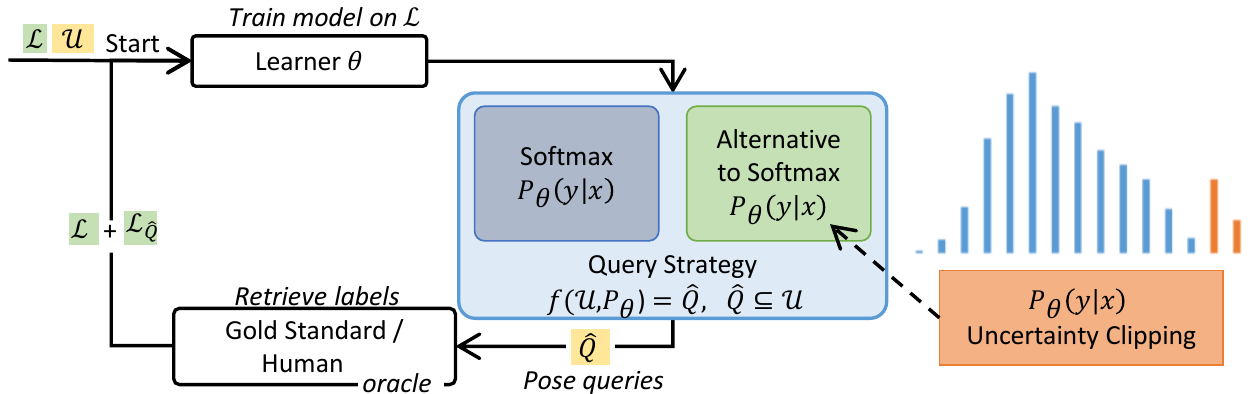}
    \caption{Standard Active Learning Cycle including our proposed \textit{\acrfull{UC}} to influence the uncertainty based ranking (using the probability $\conf_\learner(y|x)$ of the learner model $\learner$ in predicting class $y$ for a sample $x$) by ignoring the top-$k$ results}
    \label{fig:al_cycle}
\end{figure}

%In contrast to most of the literature focusing on measuring the confidence of an \gls{NN} we are not primarily interested in the best method measuring the most correct confidence, but in the effectiveness of using alternative methods as parts of an \gls{AL} strategy. We show that most alternative methods indeed work well in measuring the confidence, but not far better than the vanilla softmax function. 

%Our main contribution is \textit{\acrfull{UC}}: As we will show later, is is surprisingly harmful in the context of \gls{AL} for Transformer models to pureley select the most uncertain samples. Those samples, which are selected for labeling by an \gls{AL} strategy, are the most uncertain samples, and therefore, the better the measurement of the confidence is, the more likely those samples are outliers in the sample space. But simply labeling outliers, unsurprisingly results in worse performance than using a potentially mediocre softmax function. As a consequence, we propose to ignore the top-5\% of the confidence measurement, therefore discarding the now found and interfering outliers, improving nearly all evaluated methods.

The remainder of this paper is structured as follows:
In Section~\ref{sec:AL_strategies} we briefly explain \gls{AL}, the Transformer model architecture, and the softmax function. Section~\ref{sec:methods} presents the alternative confidence measurement techniques, Section~\ref{sec:impl_details} describes our experimental setup. Results are discussed in Section~\ref{sec:results}. In Section~\ref{sec:related_works} we present related work and conclude in Section~\ref{sec:conclusion}.

\section{\acrlong{AL} 101}
\label{sec:AL_strategies}
Supervised learning techniques inherently rely on an annotated dataset dataset.
\gls{AL} is a well-known technique for saving human effort by iteratively selecting exactly those unlabeled samples for expert labeling that are the most useful ones for the overall classification task.
The goal is to train a classification model $\learner$ which maps samples $x \in \mathcal{X}$ to a respective label $y \in \mathcal{Y}$; for the training, the labels $\mathcal{Y}$ have to be provided ``somehow''.
%
%Generally, there exist three widely used \gls{AL} scenarios: i) pool-based \gls{AL}, in which a large set of unlabeled samples is existent from the start and is consecutively labelled, ii) \textit{membership query synthesis}, in which the \gls{AL} strategy asks for labels for any unlabeled sample possible in the vector space - even artificially created samples which are not existent in the real-world domain, and iii) \textit{stream-based selective sampling}, in which the \gls{AL} strategy decides for a theoretically infinite stream of unlabeled samples per sample consecutively to label or to discard the sample~\cite{settles_al_survey}. We focus on the pool-based scenario, as this is the most practical one to encounter in \gls{ML} projects.
Figure~\ref{fig:al_cycle} shows a standard pool-based \gls{AL} cycle:
Given a small initial labeled dataset $\labeledSet = \{(x_i,y_i)\}_{i=0}^n$ of $n$ samples $x_i \in \mathcal{X}$ and the respective label $y_i \in \mathcal{Y}$ and a large unlabeled pool $\unlabeledSet = \{x_i\}, x_i \not\in \labeledSet$, an \gls{ML} model called \textit{learner} $\learner: \mathcal{X} \mapsto \mathcal{Y}$ is trained on the labeled set.

A \emph{query strategy} $\ALStrategy\colon \unlabeledSet \longrightarrow \unlabeledQuery$ then subsequently chooses a batch of $\batchSize$ unlabeled samples $\unlabeledQuery$, which will be labeled by the oracle (human expert) and added to the set of labeled data $\labeledSet$. This \gls{AL} cycle repeats $\tau$ times until a stopping criterion is met.

%At its core, the vast majority of \gls{AL} strategies rely on the same set of two simple heuristics: \textit{informativeness/uncertainty} and \textit{representativeness/diversity}. The first favors samples that foremost improve the model, whereas the latter prefers samples that represent the overall sample distribution in the feature vector space. Most recent \gls{AL} strategies~\cite{ALIL, CAL, BatchBALD, SPAL} aim to combine both heuristics.%, but at its core they all use the same two ingredients.

%Informativeness utilizes the classification boundary to select those samples first for labeling, which are the closest to to the boundary, whereas representativeness samples evenly distributed in the vector space. For simple \gls{ML} models like Support-Vector-Machines the boundary is explicitly defined.  As application domain we focus on \gls{NLP} and use therefore the Transformer architecture based Deep \gls{NN} models. 
%\todo{ende}

In its most basic form, the focus of the sampling criteria is on either \textit{informativeness/uncertainty}, or \textit{representativeness/diversity}. Informativeness in this context prefers samples, which reduce the error of the underlying classification model by minimizing the uncertainty of the model in predicting unknown samples.  Representativeness aims at an evenly distributed sampling in the vector space of the samples. As we are interested in improving the core measure of informativenesses, the confidence of the \gls{ML} model in its own predictions, we are concentrating in this paper on evaluating \gls{AL} strategies, which are solely relying on the informativeness metric.

Commonly used \gls{AL} query strategies, which are relying on informativeness, use the confidence of the learner model $\learner$ to select the \gls{AL} query.
The confidence is defined by the probability of the learner $\conf_{\learner}(y|x)$ in classifying a sample $x$ with the label $y$.
The most simple informativeness \gls{AL} strategy is \textit{\acrlong{LC}} (\acrshort{LC})~\cite{lc_sampling}, which selects those samples, where the learner model is most uncertain about, i.e. where the probability $\conf_{\learner}(\hat{y}|x)$ of the most probable label $\hat{y}$ is the lowest:
\begin{equation}
    \LC(\unlabeledSet) = \argmax_{\singleSample \in \unlabeledSet}\left(1-\conf_{\learner}(\hat{y}|x)\right)
\end{equation}
As we are interested in the effectiveness of alternative methods in computing the confidence probability $\conf_{\learner}$, we are using each method in combination with \acrfull{LC}, which directly relies solely on the confidence without further processing.

\section{Confidence Probability Quantification Methods}
\label{sec:methods}
The confidence probability of an \gls{ML} model should represent, how probable it is that the prediction is true. For example, a confidence of $70\%$ should mean a correct prediction in $70$ out of $100$ cases. We are using \acrfull{NN} as the \gls{ML} model. 
Due to the property of having one as the sum for all components the \textit{softmax} function is often used as a makeshift probability measure for the confidence of \glspl{NN}:
\begin{equation}
    \softmax(\logit_i) = \frac{exp(\logit_i)}{\sum^K_{j=1}exp(\logit_j)}, \text{ for } i=1,\dots, K
\end{equation}
The output of the last neurons $i$ before entering the activation functions is called \textit{logit}, and denoted as $\logit_i$, $K$ denotes the amount of neurons in the last layer.
But as has been mentioned in the past by other researchers~\cite{lakshminarayanan2017simple,weiss2022simple,gleave2022uncertainty,sankararaman2022bayesformer,rejected_openreview}, the training objective in training \glspl{NN} is purely to maximize the value of the correct output neuron, not to create a true confidence probability. An inherent limitation of the softmax function is its inability to have -- in the theoretical case -- zero confidence in its prediction, as the sum of all possible outcomes always equals 1. Previous works have indicated that softmax based confidence is often overconfident~\cite{MC_Dropout}. Especially \glspl{NN} using the often used ReLU activation function~\cite{ReLU} for the inner layers can be easily tricked into being overly confident in any prediction by simply scaling the input $\singleSample$ with an arbitrarily large value $\alpha>1$ to $\tilde{\singleSample} = \alpha\singleSample$ ~\cite{pearce2021understanding,hein2019relu}.

We selected seven methods from the literature suitable to quantify the confidence of Deep \glspl{NN} such as Transformer models. They can be divided into four categories~\cite{gawlikowski2021survey}: a) single network deterministic methods, which deterministically produce the same result for each \gls{NN} forward pass (\textit{\acrfull{IS}}~\cite{inhibited_softmax}, \textit{\acrfull{TrSc}}~\cite{trustscore} and \textit{\acrfull{Evi}}~\cite{evidential}), b) Bayesian methods, which sample from a distribution and result therefore in non-deterministic results (\textit{\acrfull{MC}}~\cite{MC_Dropout}), c) ensemble methods, which combine multiple deterministic models into a single decision (\textit{Softmax Ensemble}~\cite{qbc_sampling} ), and d) test-time augmentation methods, which, similarly to the ensemble methods, augment the input samples, and return the combined prediction for the augmented samples. The last category is a subject of future research as we could not find a subset of data augmentation techniques which reliably worked well for our use case among different datasets.

Additionally to the aforementioned categories, the existing softmax function can be \textit{calibrated} to produce meaningful confidence probabilities. For calibration we selected the two techniques \textit{\acrfull{LS}}~\cite{szegedy2016rethinking} and \textit{\acrfull{TeSc}}~\cite{zhang2020mix}.

More elaborate \gls{AL} strategies like \textit{BALD}~\cite{BatchBALD} or \textit{QUIRE}~\cite{QUIRE} not only focus on the confidence probability measure, but also make use of the vector space to label a diverse training set, including also regions far away from the classification boundary. As the focus of this paper is on purely evaluating the influence of the confidence prediction methods, we are deliberately solely using the most basic \gls{AL} strategy \acrlong{LC}.

In the following the core ideas of the individual methods are briefly explained, more details, reasonings, and the exact formulas can be found in the original papers. In the end we are explaining our meta-strategy \textit{\acrlong{UC}}, which significantly enhances several confidence probabilities quantification methods.

\paragraph{\acrlong{IS} (Single Network Deterministic Model)}
The \acrlong{IS} method~\cite{inhibited_softmax} is a simple extension of the vanilla softmax function by an additional constant factor $\alpha \in \mathbb{R}$:
\begin{equation}
    \softmax(\logit_i) = \frac{exp(\logit_i)}{\sum^K_{j=1}exp(\logit_j)+exp(\alpha)}
\end{equation}
The constant factor enhances the effect of the absolute magnitude of the single logit value $\logit_i$ on the softmax output. To ensure that the added fraction is not removed during the training process, several changes to the \gls{NN} have to be made, including: a) removing the bias $b$ from the input of the neuron activation function, and b) extending the loss function by a special \textit{evident} regularisation term.

\paragraph{\acrlong{TrSc} (Single Network Deterministic Model)}
The \acrlong{TrSc}~\cite{trustscore} method uses the set of available labeled data to calculate a \textit{\acrlong{TrSc}}, independent on the \gls{NN} model. In a first step the available labeled data is clustered into a single high density region for each class. The \acrlong{TrSc} $ts$ of a sample $\singleSample$ is then calculated as the ratio between the distance from $\singleSample$ to the cluster of the nearest class $c_{closest}$, and the distance to the cluster of the predicted class $\hat{y}$:
\begin{equation}
    ts_\singleSample = \frac{\dist(\singleSample, c_{closest})}{\dist(\singleSample, c_{\hat{y}})}
\end{equation}
Therefore, the \acrlong{TrSc} is higher when the cluster of the nearest class is further away from the cluster of the most probable class, indicating a potentially wrong classification. The distance metric as well as the calculation of the clusters is based on the $k$-nearest neighbors algorithm.

\paragraph{\acrlong{Evi} (Single Network Deterministic Model)}
\acrlong{Evi}~\cite{evidential} treat the vanilla softmax outputs as a parameter set over a dirichlet distribution. The prediction acts as \textit{evidence} supporting the given parameter set out of the distribution, and the confidence probability of the \gls{NN} reflects the dirichlet probability density function over the possible softmax outputs.

\paragraph{\acrlong{MC} (Bayesian Method)}
\acrlong{MC}~\cite{MC_Dropout} is a Bayesian method that uses the \gls{NN} dropout regularization method to construct for ``free'' an ensemble of the same trained model. Dropout refers to randomly disabling neurons during the training phase, originally with the aim of reducing overfitting of the to-be-trained network. For \acrlong{MC}, the dropout method is applied during the prediction phase. As neurons are disabled randomly, this results in a large Gaussian sample space of different models. Therefore, each model, with differently dropped out neurons, results in a potentially different prediction. Combining the vanilla softmax prediction using the arithmetic mean produces a combined prediction confidence.

\paragraph{Softmax Ensemble (Ensemble Method)}
The softmax ensemble approach uses an ensemble of \gls{NN} models, similar to \acrlong{MC}. The predictions of the ensemble can be interpreted as a vote upon the prediction. The disagreement among the votees acts then as the confidence probability and can be calculated in two ways, either as \gls{VE}, or as \gls{KLD}~\cite{KLD}:
\begin{equation}
    VE(\singleSample) = -\sum_i\frac{V(\hat{y}, \singleSample)}{K}log\frac{V(\hat{y}, \singleSample)}{K}
\end{equation}
with $K$ being the number of ensemble models, and $V(\singleSample, \hat{y})$ denoting the number of ensemble models assigning the class $\hat{y}$ to the sample $\singleSample$. The complete equation to calculate the \gls{KLD} is omitted for brevity. Using an ensemble of softmax models inside of an \gls{AL} strategy results in the \gls{QBC}~\cite{qbc_sampling} strategy.

\paragraph{\acrlong{TeSc} (Softmax Calibration Method)}
\acrlong{TeSc}~\cite{zhang2020mix} is a model calibration method that is applied after the training and changes the calculation of the softmax function by inducing a temperature $T>0$:
\begin{equation}
    \softmax(\logit_i) = \frac{exp(\logit_i/T)}{\sum^K_{j=1}exp(\logit_j/T)}
\end{equation}
For $T=1$ the softmax function stays the same as the original version, for values $T<1$ the softmax output of the largest logit is increased, and for values of $T>1$ (which is the recommended case in using \acrlong{TeSc}) the output of the most probable logit is decreased. This has a dampening effect on the overall confidence.

The value for the temperature $T$ is computed empirically using the existent labeled set of samples during application time. The parameter changes therefore for each \gls{AL} iteration.

\paragraph{\acrlong{LS} (Softmax Calibration Method)}
\acrlong{LS}~\cite{szegedy2016rethinking} removes a fraction $\alpha$ of the loss function per predicted class and distributes it uniformly among the other classes by adding $\frac{\alpha}{K-1}$ to the other loss outputs, with $K$ being the number of classification classes. In contrast to \acrlong{TeSc} \acrlong{LS} is not applied post-hoc to a trained network, but has a direct influence on the training process due to a changed loss function.

\subsection{\acrlong{UC}}

\begin{figure}[!t]
     \centering
     \begin{subfigure}[b]{0.3\textwidth}
         \centering
         \includegraphics{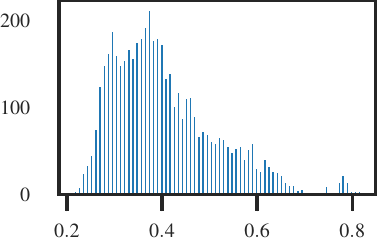}
         \caption{\acrlong{LS}}
     \end{subfigure}
     \hfill
     \begin{subfigure}[b]{0.3\textwidth}
         \centering
          \includegraphics{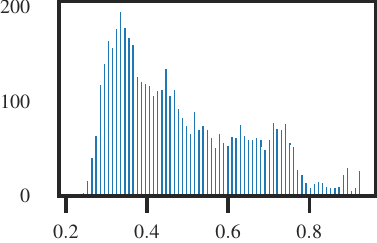}
         \caption{\acrlong{TrSc}}
     \end{subfigure}
     \hfill
     \begin{subfigure}[b]{0.3\textwidth}
         \centering
          \includegraphics{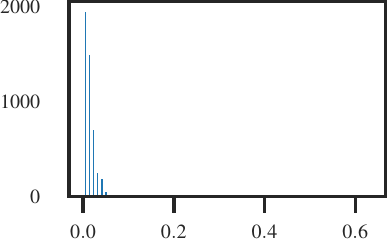}
         \caption{Full Passive Classifier}
     \end{subfigure}
     \hfill
         \caption{Exemplary uncertainty values (equals one minus classification confidence probability) for single \gls{AL} iteration of TREC-6 dataset before \acrlong{UC} as histograms}
        \label{fig:unc_hists}
\end{figure}

The aforementioned methods to quantify the confidence of \glspl{NN} in their predictions can directly be used in \gls{AL} strategies to sort the pool of unlabeled samples to select exactly those samples for labeling that have the lowest confidence/highest uncertainty. As repeatedly reported by others~\cite{karamcheti-etal-2021-mind, gleave2022uncertainty, sankararaman2022bayesformer, rejected_openreview}, \gls{AL} is rarely able to outperform pure random sampling when used to fine-tune Transformer models. Purely labeling based on the uncertainty score results in labeling many outliers/hard-to-learn-from samples, which results in an often bad classification performance. 

Figure~\ref{fig:unc_hists} displays exemplarily the histograms of the prediction probabilities/uncertainty values for the two methods \acrlong{LS} and \acrlong{TrSc} for a single \gls{AL} iteration for the TREC-6 dataset. Both distributions have a characteristic small peak of uncertainty to the far right, and we theorize, that these are the outliers which are labeled first. For comparison, the uncertainty values of a \textit{passive} classifier are shown -- an \gls{NN} model trained on the full available training dataset --, such a model is obviously very confident in its predictions.

To prevent this issue we propose the na\"ive, but powerful method \textit{\acrlong{UC}}, where the top-$k\%$ of the most uncertain samples are ignored by the \gls{AL} strategy, resulting in ignoring the displayed small peaks to the far right. Therefore, not the most uncertain samples -- potentially many outliers -- but the second-most uncertain samples are labeled. This method can be used in combination with any \gls{AL} strategy using uncertainty for ranking the pool of unlabeled samples.

\section{Experimental Setup}
Details to reproduce our evaluation are provided in this chapter. We are open to join the reproducability initiative and to offer other researchers the re-use of our work by making our source code fully publicly available on GitHub~\footnote{\url{https://github.com/jgonsior/btw-softmax-clipping}}.

\label{sec:impl_details}
\subsection{Setup}
We extended the \gls{AL} framework \textit{small-text}~\cite{small-text}, tailored for the use-case of applying \gls{AL} to Transformer networks. Some of the aforementioned methods such as \acrlong{LS} can be applied directly during the training of the network and have a positive effect on the training outcome, whereas others like \acrlong{MC} are applied after the training is complete. As we are only interested in evaluating the effect of alternative confidence probability quantification methods instead of the potentially positive influence on the classification quality, we are effectively training two Transformer models simultaneously. One is the original vanilla Transformer model with a linear classifier and a softmax on top of the CLS embedding. This one is used for class predictions and for calculating the classification quality. The other one is solely used for the \gls{AL} selection process, and includes the implemented alternative confidence probability quantification methods. Even though this adds computational overhead to our experiments, it enables us to evaluate the effect of the confidence quantification for \gls{AL} independently of potentially other positive -- or negative -- effects on the classification quality.

Parameters for the compared methods were selected empirically using hyper parameter tuning. The constant factor $\alpha$ for \acrlong{IS} was set to $1.0$, for \acrlong{TrSc} the $k$ for the $k$-nearest neighbor calculation to 10. We used an ensemble of $50$ models for \acrlong{MC}, each simple softmax based Transformer model with a different seed for the random-number-generator. For the softmax ensemble we could only use 5 different Transformer models with the vanilla softmax activation function, as the runtime was drastically higher compared to \acrfull{MC}. For \acrlong{LS} the fraction $\alpha$ was set to $0.2$, and for
\acrlong{TeSc} 1.000 different values for $T$ between $0$ and $10$ were tested, the temperature resulting in the smallest cross entropy was finally used. Whenever possible, we used the original implementations of the methods with slight adaptations to make them work together with Transformer models and the small-text framework.

\paragraph{Active Learning Simulation}
\begin{table}[!t]
\centering
\small
\begin{tabular}{lr}
\toprule
 AL strategy                   & Abbreviation   \\
\midrule
 \acrlong{LC} (\acrlong{Evi})     & \acrshort{Evi}            \\
 \acrlong{LC} (\acrlong{IS})         & \acrshort{IS}             \\
 \acrlong{LC} (\acrlong{LS}) & \acrshort{LS}             \\
 \acrlong{LC} (MonteCarlo)      & \acrshort{MC}           \\
 \acrlong{LC} (Softmax Ensemble: \acrlong{KLD})    & \acrshort{KLD}            \\
 \acrlong{LC} (Softmax Ensemble: \acrlong{VE})     & \acrshort{VE}             \\
 \acrlong{LC} (\acrlong{TeSc})    & \acrshort{TeSc}           \\
 \acrlong{LC} (\acrlong{TrSc}) & \acrshort{TrSc}           \\
\midrule
 Random       & Rand           \\
 \acrlong{LC} (softmax)         & \acrshort{LC}             \\
 \acrlong{MM} (softmax)         & \acrshort{MM}             \\
 \acrlong{Ent} (softmax)        & \acrshort{Ent}            \\
\midrule
 Passive (full trained set)    & Pass           \\
\bottomrule
\end{tabular}
\caption{Abbreviation for the evaluated \gls{AL} strategies using the Confidence Probability Quantification Methods including several baselines}
\label{tab:al_strats}
\end{table}
To evaluate the effectiveness of the confidence probability methods we are simulating \gls{AL} in the following way: For a given labeled dataset, we start with an initially labeled set of 25 samples, and perform afterwards 20 iterations of \gls{AL}, ignoring the known labels for the other samples, following the procedure of \cite{Zhang_AL} and \cite{Schroeder2022}. In each iteration a batch of 25 samples is selected by the \gls{AL} strategy for labeling, which is performed using the known ground-truth labels. As \gls{AL} strategies we used the most simple strategy, directly using the confidence probability measure without further pre-processing for ranking, \acrlong{LC} (see Section~\ref{sec:AL_strategies}) in combination with the softmax function replaced with the alternative confidence probability measures. As additional baselines we included \acrlong{Ent}~\cite{ent_sampling}, \acrlong{MM}~\cite{mm_sampling} and pure random sampling. The first two further processed the confidence values before ranking by either calculating the entropy of the uncertainty, or calculating the margin between the first and second-most probable class, but did not make use of any other information. Each simulation was repeated 10 times using different initially labeled samples. Another baseline was the passively trained model on all available training data. The used strategies are summarized in Table~\ref{tab:al_strats} including their abbreviations used in the plots in the remainder of this paper.

%The underlying architecture allows to pre-train a Transformer model on a large text corpus in an unsupervised manner. After the pre-training phase the Transformer model has knowledge about natural language and can semantically model text. When applying a Transformer model in a supervised context for classification tasks, the pre-trained model can be extended by a final feed-forward \gls{NN} layer, which then gets trained on a set of labeled data (\textit{finetuning}). The most prominent pre-trained Transformer-encoder model is BERT~\cite{BERT}.

\subsection{Transformer Models}
A \textit{transformer}~\cite{transformer} is a deep \gls{NN} model using the self-attention~\cite{transformer} mechanism to process sequential input. We use as Transformer models the original $BERT_{base}$~\cite{BERT} model and the updated version $RoBERTa_{base}$~\cite{roberta}. The networks of $BERT_{base}$ and $RoBERTa_{base}$ consist of 12 layers with an embedding size of 768, with BERT having 110M parameters, and RoBERTa 125M parameters. We use the pretrained versions from Hugginface~\cite{huggingface}. As the \gls{AL} experiments require a significant amount of fine-tuning the Transformer models we deliberately use the smaller \textit{base} version instead of the \textit{large} versions of both Transformer models. Even though the large versions result in presumably better classification accuracies, the relative differences in the compared methods should be nearly the same. Each pre-trained Transformer network is extended by a final fully connected projection layer based on the sentence representation ``CLS'' token, either using the vanilla softmax function, or using the alternative methods presented in this paper.

\subsection{Datasets, Hardware, and Metrics}
\begin{table}[!t]
\centering
\small
\begin{tabular}{llcrr}
\toprule
Dataset Name (abbreviation)         & Domain    & \# classes & \#Train & \#Test \\ \midrule
AG's News (AG)      & News      & 4          & 120,000 & 7,600  \\
CoLA (Co)      & Grammar      & 2          & 8,551 & 527  \\
IMDB (IM)          & Sentiment & 2          & 25,000  & 25,000 \\
Rotten Tomatoes (RT) & Sentiment & 2           & 8,530   & 1,066  \\
Subjectivity (SU)  & Sentiment & 2          & 8,000   & 2,000  \\
SST2 (S2)  & Sentiment & 2          & 6,920   & 1,821  \\
TREC-6 (TR)      & Questions & 6          & 5,452   & 500    \\ \bottomrule
\end{tabular}
\caption{Information about the datasets used in the experiments.}
\label{tab:datasets}
\end{table}
Table~\ref{tab:datasets} lists the datasets we used in our experiments to \textit{fine-tune} the pre-trained Transformer models in the \gls{AL} simulations. The datasets were selected as a diverse set of popular \gls{NLP} datasets, including binary and multi-class classification of different domains of varying difficulty. The datasets were obtained from the Huggingface dataset repository~\cite{datasets}. We used the train-test-splits provided by Huggingface. All experiments were conducted on a cluster consisting of NVIDIA A100-SXM4 GPUs, AMD EPIC CPU 7352 (2.3GHZ) CPUs and NVMEe disks. Each experiment was run on a single graphic card, with 120GB memory, and 16 CPU cores.

The \gls{AL} experiments can be evaluated in multitude of ways. At its core, after each \gls{AL} iteration, a standard \gls{ML} metric is being measured for the labeled and withheld dedicated test set. We decided upon the \textit{accuracy (acc)} metric, calculated on a withheld test dataset. It is possible to compare the test accuracy values of the last iteration, the mean of the last five iterations ($acc_{last5}$), and the mean of all iterations. The last one equals the area-under-the-curve when plotting the so-called \gls{AL} learning curve. As an effective \gls{AL} strategy should select the most valuable samples for labeling first, those metrics that include the accuracy of multiple iterations are often closer to real use-cases. Nevertheless, at the beginning of the labeling process the fluctuation of the test accuracy for most strategies is very high and contains often surprisingly few information about which strategy is better than another one. The influence of the initially labeled samples is simply so high, that a better strategy, with a bad starting point, has no chance to be better than a bad strategy, which has a good starting point. But after a couple of \gls{AL} iterations, the results stabilize, and good strategies can be reliably distinguished from bad strategies, as each strategy tends to approach its own characteristic threshold, nevertheless the starting point. We decided therefore to use the accuracy of the last five iterations, deliberately ignoring therefore the first iterations with the highly fluctuating results.

%Each of the \gls{AL} experiments was run with a fixed budget of 20 iterations, each labeling a batch of 25 samples. Therefore, we use the final accuracy after the last labeled samples were added, computed on a withheld test dataset. As each method had the same budget of labelled samples, the final accuracy suffices to compare the result of the iterative process. An alternative approach would be take the test accuracy after each \gls{AL} iteration into account, e. g. by using the average of the accuracies. But as the quality of the intermediate decisions during the earlier \gls{AL} decisions depend highly on the initially labeled samples at the beginning of the experiments

Additionally, we measured the runtime. \gls{AL}, applied in real-life scenarios, is an interactive process. Decisions of the \gls{AL} strategy should be made in the magnitude of single-digit seconds, longer calculation times render the annotation process often unusable. %We measured only the time the \gls{AL} strategies needed for making their decisions upon which samples to label next, as the training time of the Transformer models is the same for almost all compared methods.

As the experiments were each repeated for 10 times, we display in the following the arithmetic mean for the 10 repetitions.

\section{Results}
\label{sec:results}
We compared the confidence quantification methods (See Section~\ref{sec:methods}) in many ways: Firstly, we are comparing the test accuracies $acc_{last5}$ of the last five iterations, with and without our proposed \acrlong{UC} (~\ref{sec:eva_violin}). Secondly, we analyze those samples queried by the \gls{AL} strategy for labeling to show which strategies behave similarly, indicated by similarly queried samples (Section~\ref{sec:eva_similar}. Then, we analyze the class distribution of the queried samples (Section~\ref{sec:eva_distribution}, and conclude with an analysis of the runtimes (Section~\ref{sec:eva_runtimes}). %Our analyzations can be summarized to investigate two main research questions: a) what is the impact of replacing the softmax function by an alternative \gls{NN} probability confidence measure, and b) what is the effect of our proposed \acrlong{UC} method. We aim to show why \acrlong{UC} works, and give an recommendation about which alternative confidence probability method to use instead of the softmax function.

\subsection{Test Accuracies}
\label{sec:eva_violin}

\begin{figure}[!t]
     \centering
     \begin{subfigure}[b]{\textwidth}
         \centering
         \includegraphics[width=\textwidth]{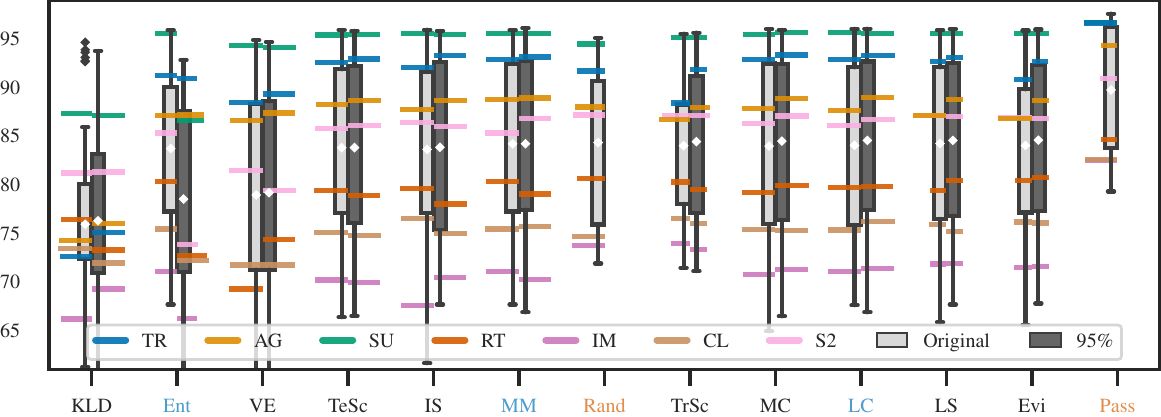}
         \caption{BERT}
     \end{subfigure}
     \hfill
     \begin{subfigure}[b]{\textwidth}
         \centering
          \includegraphics[width=\textwidth]{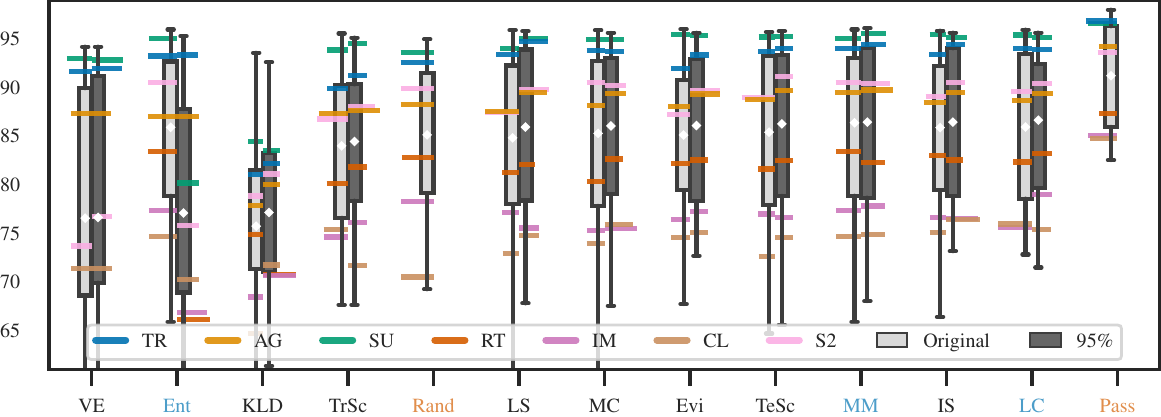}
         \caption{RoBERTa}
     \end{subfigure}
     \hfill
        \caption{Distribution of $acc_{last5}$ including the \textit{\acrlong{UC}} variants and the average $acc_{last5}$ values per dataset as colorful line, ordered by $acc_{last5}$ after \acrlong{UC}. The arithmetic mean of the runs per method are included as a white diamond in the middle of the plots. The vanilla softmax based baselines Ent, MM und LC are marked in blue, and the baselines Random Selection as well as the Passive classifier are marked orange.}
        \label{fig:violin}
\end{figure}
Figure~\ref{fig:violin} displays the distributions of the average test accuracy of the last five \gls{AL} iterations $acc_{last5}$ per method. The light grey boxplot displays the original values, the dark grey the variant with the top-5\% of the uncertainty values ignored. The underlying displayed distribution consists of the $acc_{last5}$ values combined for all datasets. As each method was evaluated using 10 different starting points, we additionally display the arithmetic mean of the 10 repetitions per dataset as an additional colorful stick. The methods are ordered after the mean $acc_{last5}$ value using \acrlong{UC}.

\paragraph{General Remarks}
First, it is obvious that the difference between the individual methods differs per dataset, but averaged over all datasets (the white diamond in the middle of the boxplots indicates the arithmetic mean) the differences become marginally small. Still, it can be safely stated already that the Softmax Ensemble techniques (\gls{KLD} and \gls{VE}) perform far worse than any other technique.

\paragraph{\acrlong{UC}}
We discarded the top-5\% of the most \textit{uncertain} values when selecting samples for labeling, displayed in the dark grey boxplots. The clipping threshold of 5\% was found empirically using hyperparameter tuning, other values up to 10\% also worked well. The reasoning behind that is that a good method for calculating confidence probabilities results in an \gls{AL} strategy querying mostly outliers, as these are the most uncertain values. But solely labeling outliers results in worse performance than Random sampling, as can be seen in Figure~\ref{fig:violin} when looking only on the light gray boxplots. Our paradoxical finding is therefore that the Softmax function is indeed a bad method for calculating the confidence probability, but purely uncertainty-based \gls{AL} strategies actually \textit{benefit} from a \textit{slightly worse-than perfect} confidence probability method.

Looking purely on the boxplots, \acrlong{UC} seems to remove some bad performing lower end results from the distributions, and add more better performing. Except for the ensemble methods and \acrlong{Ent}, all methods benefit from \acrlong{UC} (comparing the overall arithmetic mean of the distributions indicated by the white diamonds). The effect of \acrlong{UC} becomes more clear if one compares the colorful lines per dataset in each boxplot from the left and the right side. These lines indicate the arithmetic mean per dataset. Especially for the dataset AG-News and Trec-6 the clipping improves the final test accuracy drastically, as the right lines are often higher than the left ones.  Some datasets such as AG-News or Trec-6 are more influenced by \acrlong{UC}, indicating a higher percentage of outliers in these datasets in comparison to datasets such as Subjectivity.

\paragraph{Baselines}
In addition to pure \acrlong{Rand} selection, and \acrlong{LC} with the vanilla softmax function, \acrlong{Ent} and \acrlong{MM} -- both also using the vanilla softmax function -- were included in our evaluation. \acrlong{Ent} seems to be the only strategy which is highly negatively influenced by \acrlong{UC}. Apparently, the calculation of the entropy on top of the softmax negates the negative impact of outliers. The baseline strategies \acrlong{LC} and \acrlong{MM} perform better on RoBERTa than on BERT, indicating that the softmax function is better calibrated to true probabilities for RoBERTa than for the original BERT model.

\paragraph{Best overall performing method}
The results differ based on the used Transformer model. In general, no alternative method was able to be better for both models than vanilla softmax used in \acrfull{LC}. \acrfull{Evi} seems promising and performs well for both Transformer models, whereas most other strategies, which perform good for one model, perform bad for the other ones. As the differences between the methods are mostly marginal, our results do not allow to rule out \acrfull{LS}, \acrfull{MC}, \acrfull{TeSc}, or \acrfull{TrSc}, as these are all close behind. Also, the baseline \acrfull{MM} seems promising.

\subsection{Queried Samples}
\label{sec:eva_similar}
\begin{figure}[!t]
    \centering
    \includegraphics[width=\textwidth]{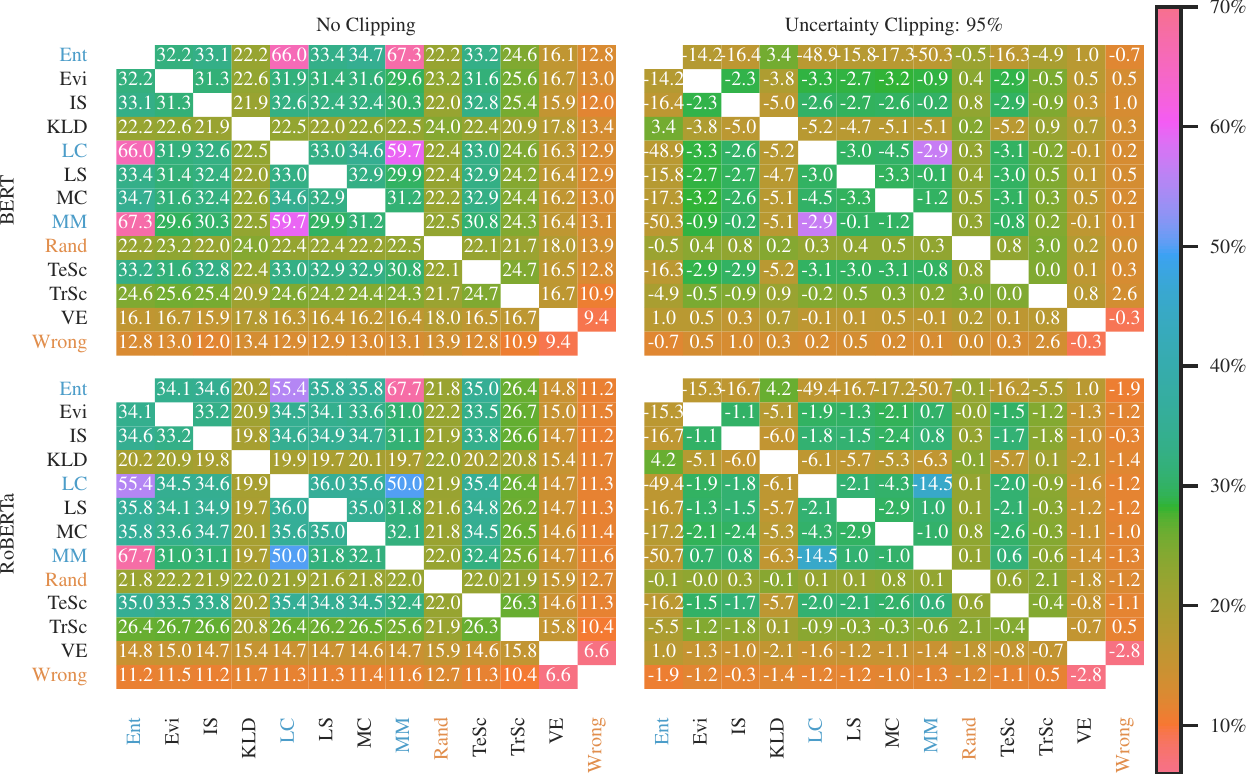}
    \caption{Heatmap for the Jaccard coefficients of the queried samples between each pair of strategies. High coefficients indicate highly similar strategies. On the right side the displayed numbers indicate the difference to the original coefficients.}
    \label{fig:queried_samples}
\end{figure}
We are also interested in the similarities and differences of the probability quantification methods. Hence, we calculated the Jaccard coefficients between the sets of queried samples for each pair of compared \gls{AL} strategies, displayed in Figure~\ref{fig:queried_samples}. As our experiments were repeated 10 times each, we used the union of the sets of queried samples, and the union of the results per dataset. Additionally, we included the wrongly classified samples (wrong) by the passive classifier -- potentially being outliers -- in the plots. The left side contains the Jaccard coefficients for the version without \acrlong{UC}, the right side with the top-5\% results being ignored. The right side contains the percentual difference to the unclipped version as numbers, whereas the color encoding still indicates the Jaccard coefficient.

For the unclipped version, the most similar strategies are unsurprisingly the three strategies using the pure softmax function: \acrlong{Ent}, \acrlong{LC}, and \acrlong{MM}. As the clipping has a high negative impact on \acrlong{Ent}, it becomes highly dissimilar from almost any other strategy, with a drop of over 40\% compared to \acrlong{LC}.

Apart from this,  \acrlong{Evi}, \acrlong{LS}, \acrlong{MC}, \acrlong{IS}, and \acrlong{TeSc} are equally similar to each other in the area of 30\%, as has been already been indicated by the similar performance in the previous section. \acrlong{TrSc} is only similar in the area of 25\%, and the Softmax Ensemble methods \acrlong{VE} and \acrlong{KLD} are highly dissimilar to everything else. Also, almost all strategies are, as expected, very dissimilar to pure random selection, and even less similar to the set of wrongly classified samples.

\acrlong{UC} seems to make all strategies behave less similar, as nearly all Jaccard coefficients go slightly down. This indicates that all strategies without \acrlong{UC} label the same set of outliers, and after these are removed, they have of course less similarities. We also included the wrongly classified samples (Wrong) by the fully trained passive classifiers in the analysis. 
Interestingly, a difference can be seen between the two compared Transformer models: for RoBERTa most strategies sample few wrongly classified samples, which becomes even less after the \acrlong{UC}, whereas for the original BERT model the percentage of queried samples increases slightly with \acrlong{UC}.

Also, all strategies become slightly more similar to random selection, which is expected, as \acrlong{UC} removes the best results based on the uncertainty heuristic, leaving the strategies labeling the second-best-results, which are often closer to pure random selection, as the best bets based on the heuristics are removed.

In conclusion, equally good performing strategies also query similar samples, and \acrlong{UC} manages to remove a set of outliers, which the different methods apparently have in common.
%\todo{übergang fehlt}
\subsection{Class distribution}
\label{sec:eva_distribution}
\begin{figure}[!t]
    \centering
    \includegraphics[width=\textwidth]{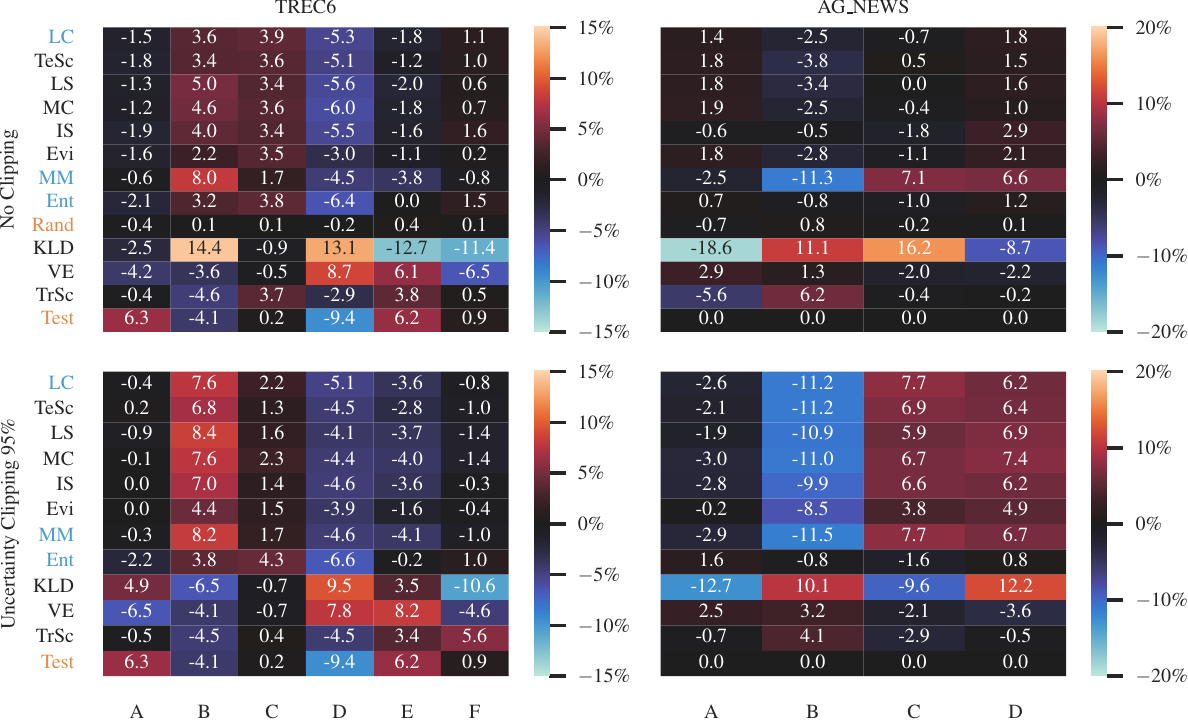}
    \caption{Difference of class distribution of the queried samples compared to the train set for the two datasets TREC-6 and AG's News for the original BERT model}
    \label{fig:class_distribution}
\end{figure}

To further analyze \acrlong{UC}, we selected the two datasets, TREC-6 and AG's News, which have more than two target classes, for an additional deeper analysis of difference in class distribution for the queried samples, compared to distribution in the training set, displayed in Figure~\ref{fig:class_distribution}. Additionally, we include the class distribution across the test dataset. The expectation of the random strategy would be the same class distribution in the queried samples as in the full available training dataset, which nearly holds true.

For TREC-6, most strategies favor samples from class B and C, and sample less samples from class D. \acrlong{UC} simply enhances this behavior. For AG's News almost all strategies sample evenly distributed among all classes without \acrlong{UC}, and significantly less often of class B, but more of class C and D with \acrlong{UC}. \acrlong{MM} stands out from the other methods, as it behaves without clipping similar to all other methods that use clipping. This explains why \acrlong{MM} is one of the few methods that does not benefit from \acrlong{UC} for Trec-6 and AG's News, as seen in Figure~\ref{fig:violin}, despite an otherwise good performance.

We infer from the data that \acrlong{UC} vastly influences the distribution of the queried classes towards potentially more interesting classes for labeling, which are the same for all methods.

\subsection{Runtime Comparison}
\label{sec:eva_runtimes}
\begin{figure}[!t]
     \centering
     \begin{subfigure}[b]{0.49\textwidth}
         \centering
         \includegraphics{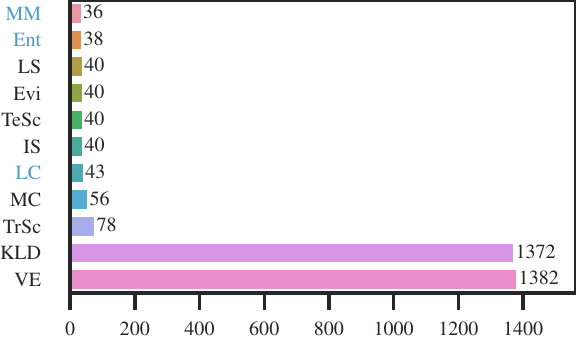}
         \caption{BERT}
     \end{subfigure}
     \hfill
     \begin{subfigure}[b]{0.49\textwidth}
         \centering
          \includegraphics{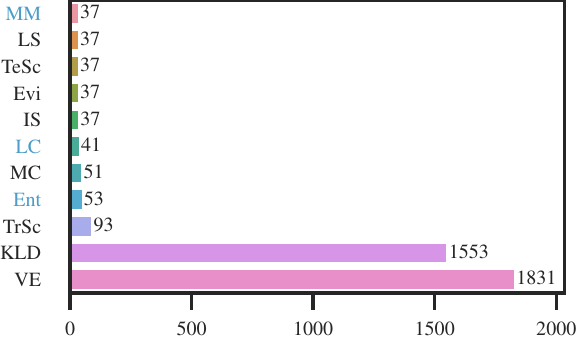}
         \caption{RoBERTa}
     \end{subfigure}
     \hfill
        \caption{Runtime Comparison of all methods averaged over the datasets in seconds}
        \label{fig:runtime}
\end{figure}

\gls{AL} is in practice an iterative process. Annotators want a responsive systems, which tells them immediately what to label next, without long waiting times. Therefore, a fast runtime of the \gls{AL} query selection is a crucial factor in making \gls{AL} real-world usable. We display in Figure~\ref{fig:runtime} the runtimes averaged over all datasets for our methods in seconds per complete \gls{AL} loop, measuring only the time of the \gls{AL} computations, not the Transformer model fine-tuning time~\footnote{The average training time of the Transformer models is around 600 seconds combined for a single \gls{AL} experiment.}. First, it becomes clear that the overhead of our Softmax Ensemble methods of training multiple Transformer models in parallel is simply too high. Apart from that, almost all other methods seem to perform equally fast, with \acrlong{MC} and \acrlong{TrSc} as the slightly, but probably neglectable, slowest methods.

\section{Related Work}
\label{sec:related_works}
The importance of a good quantification method for prediction confidence probabilities for \gls{AL} has been noted often before~\cite{lakshminarayanan2017simple,blundell2015weight}, especially in the context of deep \gls{NN} like Transformer models~\cite{Schroeder2022}. Nevertheless, few research has been done on solely comparing certainty probability methods, with the goal of \gls{AL} for Transformer models in mind. 

\cite{weiss2022simple} compared \acrlong{MM}, \acrlong{Ent}, \acrlong{MC}, as well as DeepGini, a modified softmax version with a fast runtime in mind. The primary focus of their work is on \textit{test input prioritizers} to work with very large datasets. They conclude that \acrlong{MC} performs better than the other strategies, but not so much that it justifies its usage over the vanilla softmax, and that more research in that area is necessary.
In \cite{gleave2022uncertainty} the usefulness of softmax ensemble methods for Transformer models was investigated, with the same conclusion as ours: ensemble methods perform far worse than even random sampling for Transformer models. The authors of  \cite{sankararaman2022bayesformer} propose an improved variant to standard \acrlong{MC}, directly extending the Transformer architecture. They also note that it is sometimes better to \textit{throw away} some labels to improve the accuracy, indicating similar observed results, which led us to the idea of \acrlong{UC}.
On a similar note, \cite{rejected_openreview} find that \gls{AL} methods, when applied to Transformer models, often perform inconsistent due to labeling too many \textit{unlearnable} outliers. Their solution to the problem is training multiple models, resulting in an ensemble approach. Our runtime comparison indicated that ensemble approaches should be used with caution due to very high runtimes, compared to all other methods. In \cite{karamcheti-etal-2021-mind} the authors also noted the highly negative impact of outliers as hard-to-learn-from samples for Transformer models through an extensive ablation study, and propose future research develop methods, to ignore these outliers.

\cite{pearce2021understanding} deeply investigates the foundations of the vanilla softmax function as confidence probability. They aim to explain why it functions surprisingly and state that the softmax function \textit{might have a sounder basis than widely believed}. Our results support this idea as \acrlong{LC} was not really outperformed by any other method.

\section{Conclusion}
\label{sec:conclusion}
Noting the importance of \gls{NN} prediction confidence probability measures, and the potential shortcomings of simply using vanilla softmax as such a measure, we experimentally compared eight alternative methods over seven datasets using both the original BERT Transformer model as well as the improved RoBERTa variant.
After discovering that better prediction confidence methods result in selecting only outliers for labeling, we proposed \acrlong{UC}, improving all methods, including vanilla softmax, which was not outperformed by any proposed method. Our evaluation focused on comparing the set of queried samples, the class distributions, as well as a runtime comparison, enabling us to better understand what \acrlong{UC} actually changes.

In conclusion, we have not found any evidence preventing us from recommending the vanilla softmax method. Still, the compared methods such as \acrlong{LS}, \acrlong{Evi}, and \acrlong{MC} are not far behind. Our proposed \acrlong{UC} improves both vanilla softmax as well as alternative methods, and should be used in practice.

Future research should compare the effectiveness of \acrlong{UC} in combination with more advance \gls{AL} strategies, which also make use of the vector space.

\section*{Acknowledgements}
This research and development project is funded by the German Federal Ministry of Education and Research (BMBF) and the European Social Funds (ESF) within the ``Innovations for Tomorrow's Production, Services, and Work'' Program (funding number 02L18B561) and implemented by the Project Management Agency Karlsruhe (PTKA). The author is responsible for the content of this publication.
    
The authors are grateful to the Center for Information Services and High Performance Computing [Zentrum für Informationsdienste und Hochleistungsrechnen (ZIH)] at TU Dresden for providing its facilities for high throughput calculations.

\printbibliography

\end{document}